\documentclass[10pt,twocolumn,letterpaper]{article}

\usepackage[pagenumbers]{iccv} 

\definecolor{iccvblue}{rgb}{0.21,0.49,0.74}
\usepackage[pagebackref,breaklinks,colorlinks,allcolors=iccvblue]{hyperref}

\def\paperID{10916} 
\def\confName{ICCV}
\def\confYear{2025}

\title{DeepMesh: Auto-Regressive Artist-mesh Creation with Reinforcement Learning}

\author{
  \textbf{Ruowen Zhao}$^{1,3}$\thanks{Equal contribution} \quad
  \textbf{Junliang Ye}$^{1,3}$\footnotemark[1]\quad
  \textbf{Zhengyi Wang}$^{1,3}$\footnotemark[1]  \\
  \textbf{Guangce Liu}$^{3}$ \quad
  \textbf{Yiwen Chen}$^{2}$ \quad
  \textbf{Yikai Wang}$^{1}$ \quad 
  \textbf{Jun Zhu}$^{1,3}$\thanks{Corresponding author.} \quad \\
  Tsinghua University$^{1}$\quad Nanyang Technological University$^{2}$\quad ShengShu $^{3}$ \\
  \url{https://zhaorw02.github.io/DeepMesh/}
}

\usepackage{float}

\begin{document}

\twocolumn[{
    \renewcommand\twocolumn[1][]{#1}
    \maketitle
    \begin{center}
    \centering
    \vspace{-20pt}
    \includegraphics[width=\linewidth]{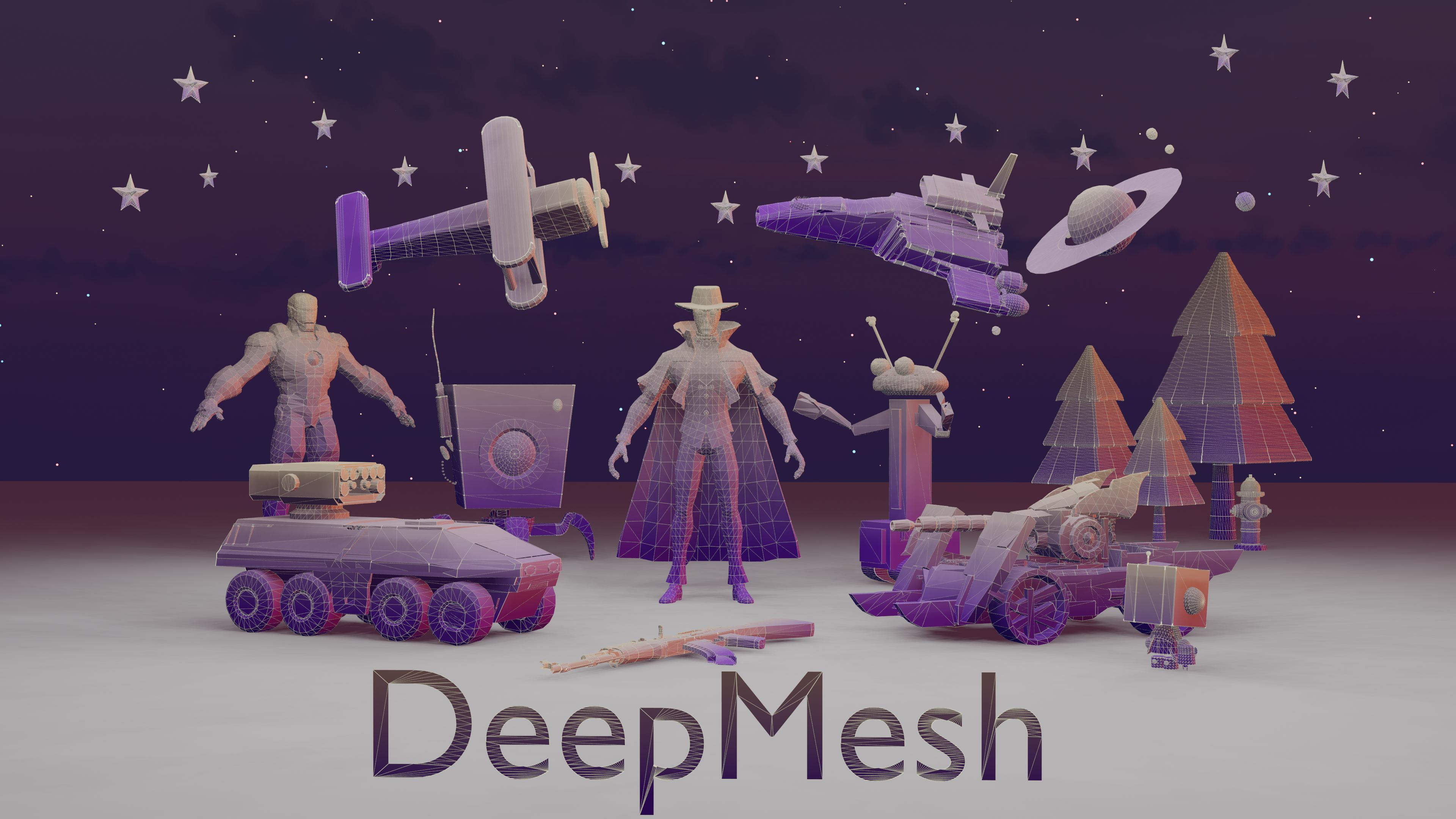}
    \captionof{figure}[Short caption]{\textbf{Gallery of DeepMesh's generation results.} DeepMesh efficiently generates aesthetic, artist-like meshes conditioned on the given point cloud.}
    \label{fig:teaser-top}
    \end{center}
}]
\newcommand\blfootnote[1]{%
\begingroup
\renewcommand\thefootnote{}\footnote{#1}%
\addtocounter{footnote}{-1}%
\endgroup
}
\blfootnote{* \ Equal contribution.}\blfootnote{\dag \ Corresponding authors.}

\begin{abstract}

Triangle meshes play a crucial role in 3D applications for efficient manipulation and rendering. While auto-regressive methods generate structured meshes by predicting discrete vertex tokens, they are often constrained by limited face counts and mesh incompleteness. To address these challenges, we propose DeepMesh, a framework that optimizes mesh generation through two key innovations: (1) an efficient pre-training strategy incorporating a novel tokenization algorithm, along with improvements in data curation and processing, and (2) the introduction of Reinforcement Learning (RL) into 3D mesh generation to achieve human preference alignment via Direct Preference Optimization (DPO). We design a scoring standard that combines human evaluation with 3D metrics to collect preference pairs for DPO, ensuring both visual appeal and geometric accuracy. Conditioned on point clouds and images, DeepMesh generates meshes with intricate details and precise topology, outperforming state-of-the-art methods in both precision and quality.

\end{abstract}

\section{Introduction}

Triangle meshes are a fundamental representation for 3D assets and are widely used across various industries, including virtual reality, gaming, and animation. These meshes can be either manually created by artists or automatically generated by applying Marching Cubes~\cite{lorensen1998marching} to volumetric fields, such as Neural Radiance Fields (NeRF) \cite{mildenhall2021nerf} or Signed Distance Fields (SDF) \cite{park2019deepsdf}. Artist-crafted meshes typically exhibit well-optimized topology, which facilitates editing, deformation, and texture mapping. In contrast, meshes generated by Marching Cubes~\cite{lorensen1998marching} prioritize geometric accuracy but often lack optimal topology, resulting in overly dense and irregular structures.

Recently, several approaches \cite{siddiqui2024meshgpt, chen2024meshanything, chen2024meshanythingv2, chen2025meshxl, tang2024edgerunner, weng2024scaling, hao2024meshtron} have emerged to generate artist-like topology from a given geometry. By taking point clouds extracted from the geometry as input, these methods learn to auto-regressively predict mesh vertices and faces, effectively preserving the structured and artistically optimized topology.

Auto-regressive mesh generation methods face two significant challenges: (1) Pre-training involves several difficulties. Tokenizing 3D meshes for transformers often leads to excessively long sequences, which increase computational costs. Moreover, training stability is further compromised by low-quality meshes with poor geometry, resulting in spikes in loss. (2) Existing methods lack mechanisms to align outputs with human preferences, limiting their ability to produce artistically refined meshes. Additionally, generated meshes often exhibit geometric defects, such as holes, missing parts, and redundant structures.

In this paper, we aim to propose a more refined and effective pre-training framework for auto-regressive mesh generation. To enhance training efficiency, we introduce an improved mesh tokenization algorithm that reduces the sequence length by 72\% without losing geometry details, which greatly reduce the training computation cost. In addition, we propose a specially designed data packaging strategy that accelerates data loading and ensures better load balancing during training. Additionally, to ensure the quality of training data, we develop a data curation strategy that filters out meshes with poor geometry and chaotic structures. This approach effectively mitigates loss spiking and enhances training stability. With these improvements, we successfully pre-train a series of large-scale transformer models for topology generation, scaling from 500 million to 1 billion parameters.

To further enhance the ability of pre-trained topology generation model, we pioneer to adapt Direct Preference Optimization (DPO)~\cite{rafailov2023direct} for 3D auto-regressive models, aligning the model outputs with human preference. First, we generate pairwise training data using the pre-trained model and annotate them with human evaluations and 3D geometry metrics. We subsequently employ reinforcement learning (RL) to fine-tune the model with these preference-labeled samples. These improvements enable our framework to generate diverse, high-quality artist-like meshes with up to 30k faces at a quantization resolution of 512.

In summary, our contributions are as follows:

1. We propose more refine pre-training framework including an efficient tokenization algorithm for high-resolution meshes along with some pre-train strategies for the auto-regressive model to facilitate efficient training. 

2. We poineer to adpat DPO to enhance our artist-mesh generative auto-regressive model with human feedback. 

\begin{figure*}[th] \label{fig:framework}
    \centering
    \includegraphics[width=\linewidth]{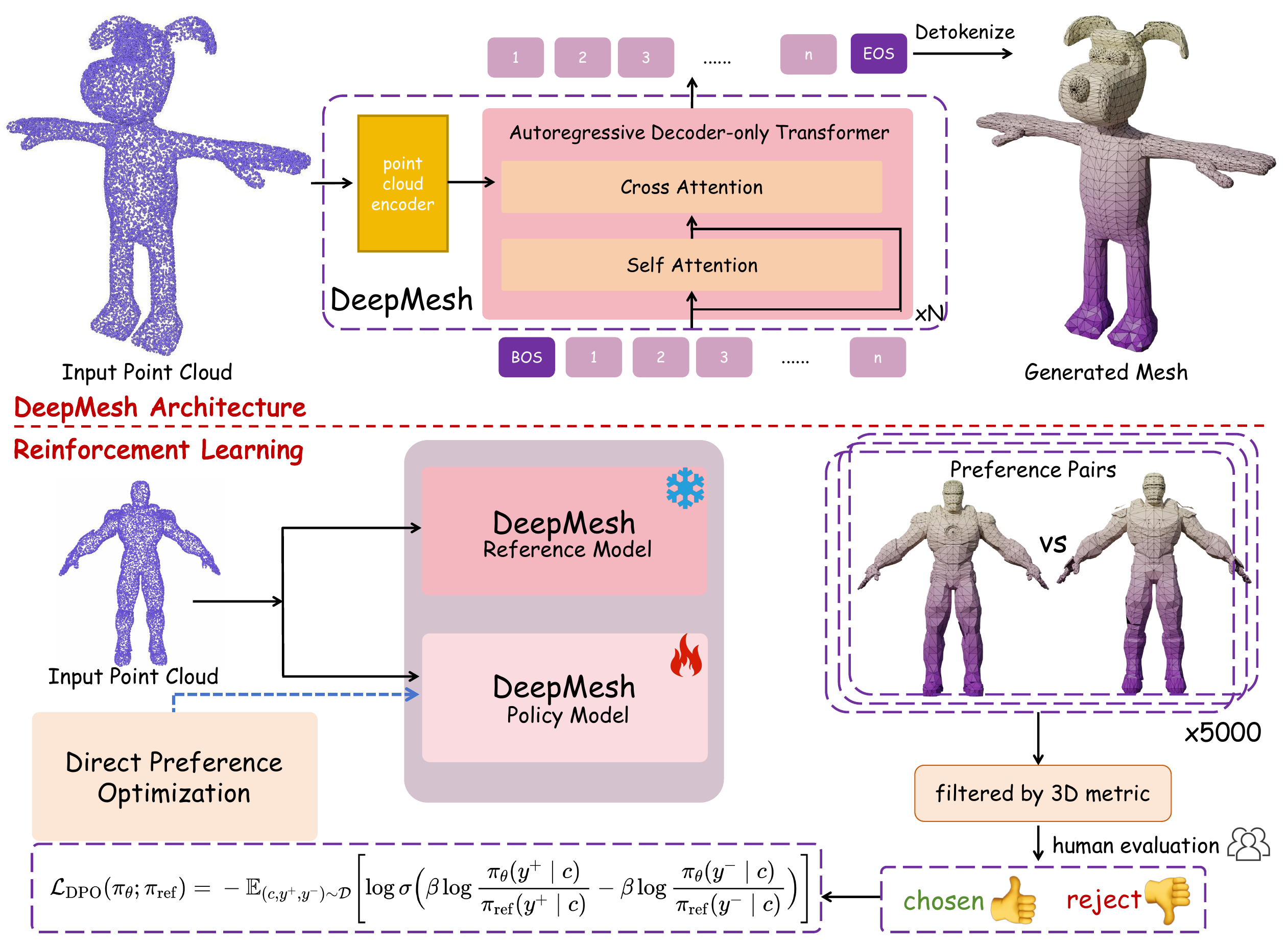}
    \vspace{-0.03\textheight}

    \caption{\textbf{An overview of our method.} DeepMesh is an auto-regressive transformer composed of both self-attention and cross-attention layers. The model is pre-trained on discrete mesh tokens generated by our improved tokenization algorithm. To further enhance the quality of results, we propose a scoring standard that combines 3D metrics with human evaluation. With this standard, we annotate 5,000 preference pairs and then post-train the model with DPO to align its outputs with human preferences. }
    \label{fig:pipeline}
    
\end{figure*}

\section{Related Work}

\subsection{3D Mesh Generation}

Early 3D generation methods utilize SDS-based optimization ~\cite{poole2022dreamfusion,wang2023prolificdreamer,sjc,lin2023magic3d,chen2023fantasia3d,sweetdreamer,raj2023dreambooth3d,chen2024text,sun2023dreamcraft3d,10.1007/978-3-031-72698-9_19,tang2023dreamgaussian,yi2024gaussiandreamer,ma2024scaledreamer} due to the limited 3D data. To tackle the Janus problem, ~\cite{shi2023mvdream,wang2023imagedream,qiu2024richdreamer,ye2024dreamreward} strengthen the semantics of different views when generating multi-view images. To minimize generation time, some approaches~\cite{zhao2024flexidreamer,long2024wonder3d,liu2023syncdreamer,liu2023one,liu2023zero,shi2023zero123++,weng2023consistent123,wu2024unique3d,voleti2024sv3d,chen2024v3d,yang2024hi3d,jiang2025animate3d} predict multi-view images and use reconstruction algorithms to produce 3D models. The Large Reconstruction Model (LRM) ~\cite{hong2023lrm} proposes a transformer-based reconstruction model to predict NeRF representation~\cite{mildenhall2021nerf} from single image within seconds. Subsequent research ~\cite{tang2024lgm,wang2024crm,xu2024grm,ziwen2024long,li2023instant3d,xu2023dmv3d,wang2023pf,NEURIPS2024_123cfe7d,zhang2024geolrm,zhang2024gs,zou2024triplane} further improve LRM's generation quality by incorporating multi-view images or other 3D representations ~\cite{kerbl20233d}. Additionally, analogous to 2D diffusion models, some early approaches~\cite{jun2023shap,nichol2022point,liu2023meshdiffusion,gao2022get3d} rely on uncompressed 3D representations, such as point clouds, to develop 3D-native diffusion models. However, these methods are often limited by small-scale datasets and struggle with generalization. More recent approaches ~\cite{wang2023rodin,zhao2023michelangelo,zhang2024clay,xiang2024structured,yang2024hunyuan3d,wu2024direct3d,li2024craftsman,chen20243dtopia,huang2025spar3d,zhang20233dshape2vecset} have focused on adapting latent diffusion models, which train a VAE to compress 3D representations. 

\subsection{Artist-like Mesh Generation}
However, All of the aforementioned works first generate 3D assets and subsequently convert them into dense meshes through mesh extraction such as Marching Cubes~\cite{lorensen1998marching}. Consequently, they fail to model the mesh topology, leading to inefficient topology such as poorly structured or tangled wireframe. Recently, approaches using auto-regressive models to generate meshes have gained attention. A pioneer work, MeshGPT~\cite{siddiqui2024meshgpt}  introduces a combination of VQ-VAE~\cite{van2017neural} and an auto-regressive transformer architecture. Subsequent works ~\cite{chen2024meshanything,weng2024pivotmesh,chen2025meshxl,wang2024llama,hao2024meshtron} explore different model architectures and extend this approach to conditional generation. However, due to the low quality of VQ-VAE, researchers propose to develop mesh quantization methods to serialize meshes. For example,  LLaMA-Mesh~\cite{wang2024llama} enables LLMs to generate 3D meshes from text prompts. MeshAnythingv2~\cite{chen2024meshanythingv2} employs Adjacent Mesh Tokenization, EdgeRunner~\cite{tang2024edgerunner} utilizes an algorithm derived from EdgeBreaker, and BPT~\cite{weng2024scaling} introduces its patchified and blocked strategy. Despite these advancements, these tokenization techniques face challenges in balancing compression ratio and vocabulary size, limiting their scalability to generate high-resolution meshes.

\subsection{RLHF with Direct Preference Optimization}
The above methods often typically adopt auto-regressive model architectures from existing large language models. With the rapid advancement of LLMs, aligning policy models with human preferences has become increasingly critical. Reinforcement Learning from Human Feedback (RLHF) is one of the most widely used post-training methods on large language models to better reflect user intentions ~\cite{yuan2023rrhf}. RLHF contains a reward model, which is trained on win-lose pairs annotated by humans, and aligns the policy model with reinforcement learning algorithms ~\cite{menick2022teaching,ouyang2022training}. However, the two-stage training pipeline often suffers from instability and imposes high computational demands. Therefore, Direct Preference Optimization (DPO) ~\cite{rafailov2023direct} has emerged as a reward model-free approach that can be easily performed. Despite DPO-based methods being extensively tested on LLMs~\cite{she2024mapo, liu2024enhancing} and VLLMs~\cite{zhou2024aligning, zhao2023beyond, li2023silkie} across text and image modalities, their application to LLMs in the 3D mesh modality remains largely unexplored.

\section{Method}
In this section, we detail our design of DeepMesh's framework. In section \ref{sec:tokenization}, we explain our improved mesh tokenization algorithm, which efficiently discretizes meshes at a high resolution and achieves an approximate 72\% compression ratio without losing geometric details. Section \ref{sec:pre-train} outlines the details of our pre-training process, including data curation, packaging and truncated training strategy. Furthermore, to enhance generation quality and align outputs with human preferences, we construct a dataset of preference pairs and post-train the model with Direct Preference Optimization (DPO)~\cite{rafailov2023direct}, as illustrated in Section \ref{sec:dpo}.

\subsection{Tokenization Algorithm} \label{sec:tokenization}
Analogous to text, meshes must be converted into discrete tokens to be processed by an auto-regressive model. In existing mesh tokenization scheme, continuous vertex coordinates are quantized into bins with a spatial resolution of $r$ and then classified into $r$ categories. After quantization, a triangular mesh is then treated as a sequence of faces, each with three discretized 3D vertex coordinates. However, this vanilla representation causes each vertex to appear as many times as the number of its connected faces, leading to considerable redundancy. Although prior works~\cite{chen2024meshanythingv2,tang2024edgerunner} have introduced tokenization methods to compress mesh sequences,  they still suffer from  relatively long token sequences, leading to increased computational costs. Recently, BPT~\cite{weng2024scaling} proposed a compressive mesh representation with a state-of-the-art compression ratio of around 74\% at 128 resolution by a local-aware face traversal and block-index coordinates encoding. However, BPT only works effectively for low-resolution meshes due to its dramatic vocabulary increase at higher resolution,  resulting in training difficulty and costs. To address these limitations, we improve its block-wise indexing to better handle high-resolution meshes.

Similar to BPT, we first traverse mesh faces by dividing them into local patches according to their connectivity to minimize redundancy in the vanilla representation. This localized traversal ensures each face only relies on a short context of previous faces, thereby avoiding long-range dependency between face tokens and mitigating the difficulty of learning. Then we sort and quantize the coordinates of each vertex in faces, and flatten them in $XYZ$ order to form a complete token sequence.   To further shorten the sequence length, we partition the whole coordinate system into three hierarchical levels of blocks and index the quantized coordinates as offsets within each block. As the quantized coordinates are sorted, neighbor vertices often share the same offset index. Therefore, we merge the indexes with the identical values to save more length. We provide more details in the supplementary material.

With these designs, our enhanced algorithm reaches approximately 72\% compression, significantly reducing sequence length and making it easier to train on high-poly datasets. Moreover, we also achieve a much smaller vocabulary size for model to learn, which improves training efficiency a lot (details are seen in Sec \ref{exp:token}).

\begin{figure}[t]
    \centering
    \includegraphics[width=\linewidth]{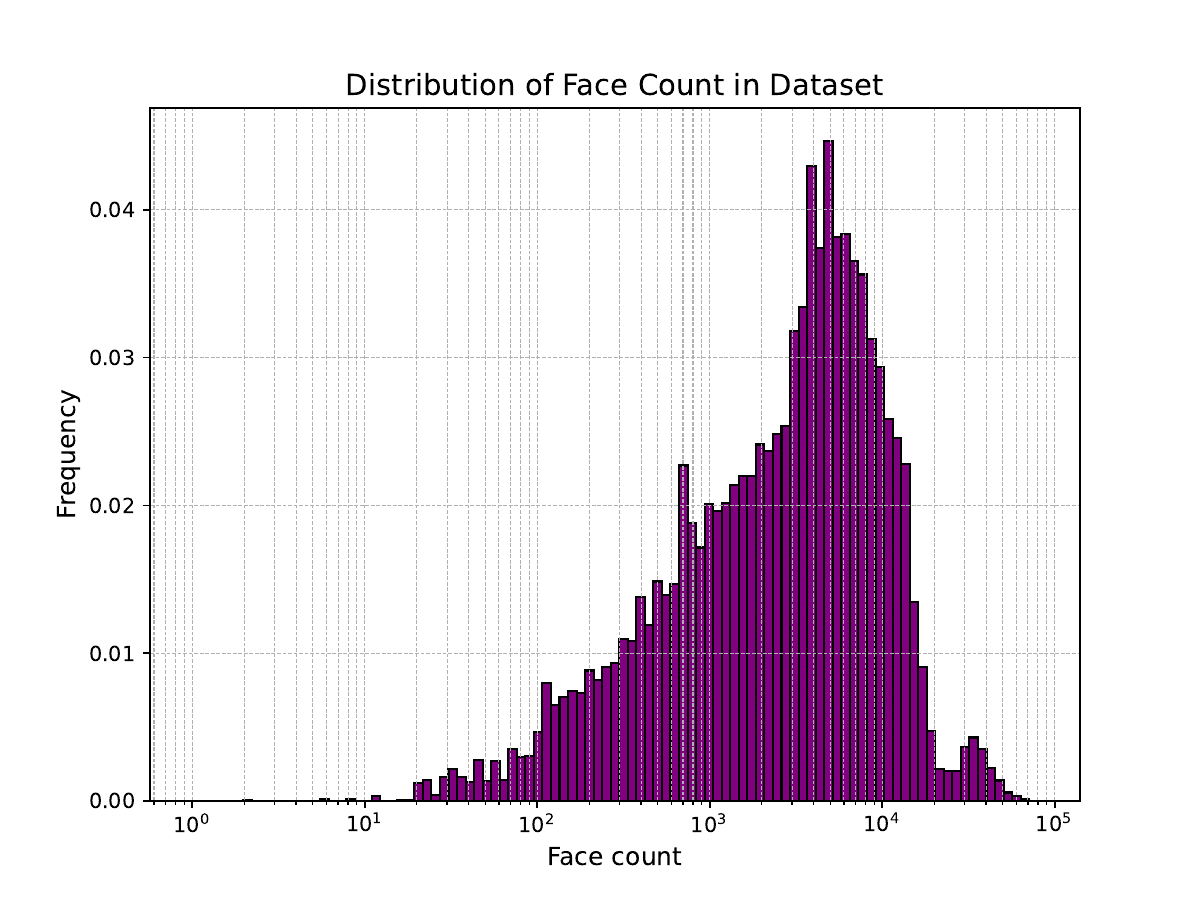}
      \vspace{-0.03\textheight}
    \caption{\textbf{Distribution of face count in training dataset.} We present the distribution of face counts in our training dataset. Our dataset size is approximately 500k, with an average face count of 8k.}
    \label{fig:faces}
\end{figure}
\begin{figure*}[th]
    \centering
    \includegraphics[width=\linewidth]{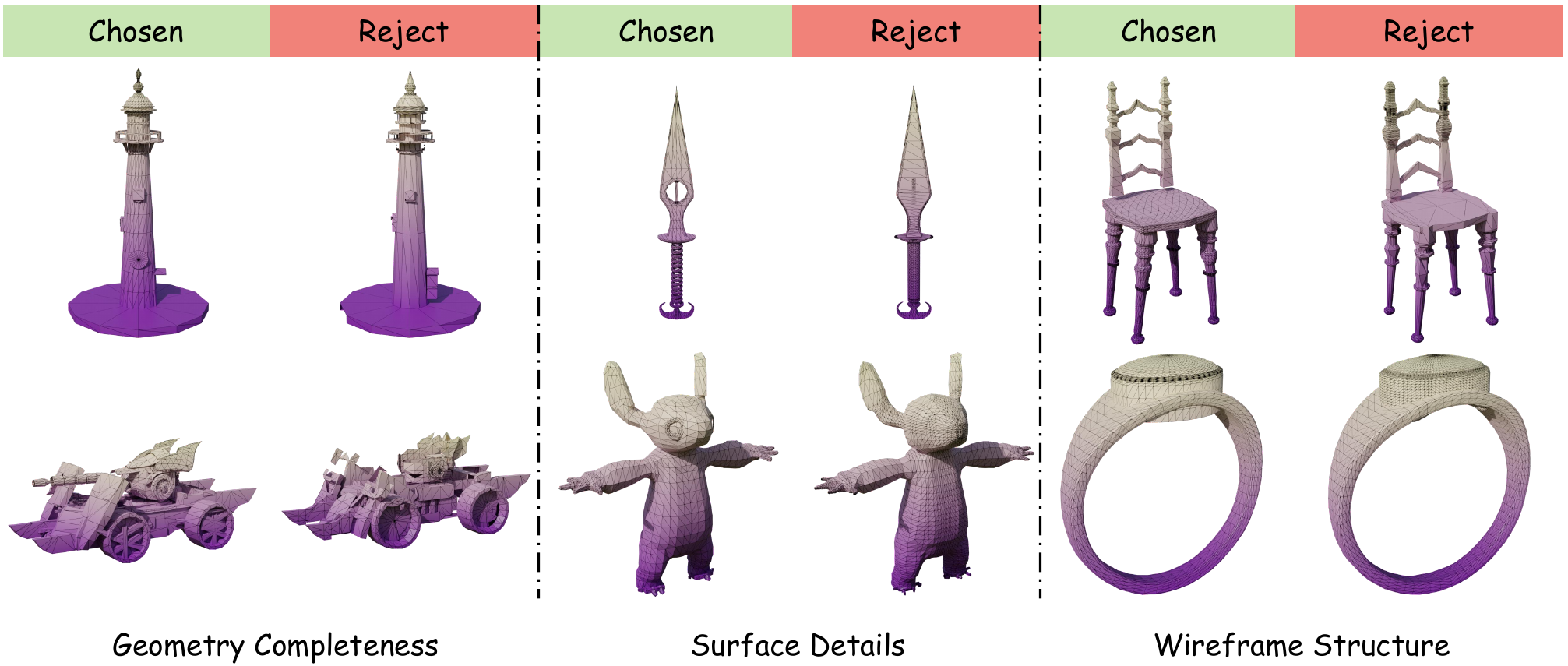}
    \vspace{-0.03\textheight}
    \caption{\textbf{Some examples of the collected preference pairs.} We annotate the preferred meshes based on their geometry completeness, surface details and wireframe structure.}
    \label{fig:dpo_data}

\end{figure*}
\subsection{Pre-training of DeepMesh} \label{sec:pre-train}
\subsubsection{Data Curation}
The quality of training data fundamentally governs model performance. However, existing 3D datasets exhibit high variability in quality, with many samples containing irregular topology, excessive fragmentation, or extreme geometric complexity. To mitigate this issue, we propose a data curation strategy that filters out poor-quality meshes based on their geometric structure and visual fidelity (more details are in supplementary material). As shown in Figure ~\ref{fig:faces}, the face count distribution of our curated dataset highlights a high-quality mesh collection.

\subsubsection{Truncated Training and Data Packaging}
As illustrated in Figure \ref{fig:faces}, high-poly meshes are prevalent in our dataset, resulting in long token sequences that significantly increase computational costs during training. To address this, we adopt truncated training from~\cite{hao2024meshtron} to enhance efficiency. Specifically,  the input token sequence is first partitioned into fixed-size context windows, with padding applied to insufficient-length segment. Then, we utilize a sliding window mechanism to shift the window step by step and train each windowed segment accordingly. To reduce unnecessary sliding in the truncated training caused by discrepancies in sequence lengths within each batch, we categorize training meshes based on face count and allocate meshes with similar face counts to each batch on each GPU. This strategy can ensure better load balancing and reduce redundant computation during training.

\subsubsection{Model Architecture}

The core structure of our DeepMesh is an auto-regressive  transformer, where each layer contains a cross-attention layer and a self-attention layer with a feed-forward network. For point cloud-conditioned generation, we employ a jointly-trained perceiver encoder based on  Michelangelo~\cite{zhao2023michelangelo}. Then the conditioned point cloud features are integrated through cross-attention. To accelerate training, we adopt the Hourglass Transformer from ~\cite{hao2024meshtron,nawrot2021hierarchical}, which can save 50\% memory while maintaining the performance.

\begin{figure*}[t]
    \centering
    \includegraphics[width=\linewidth]{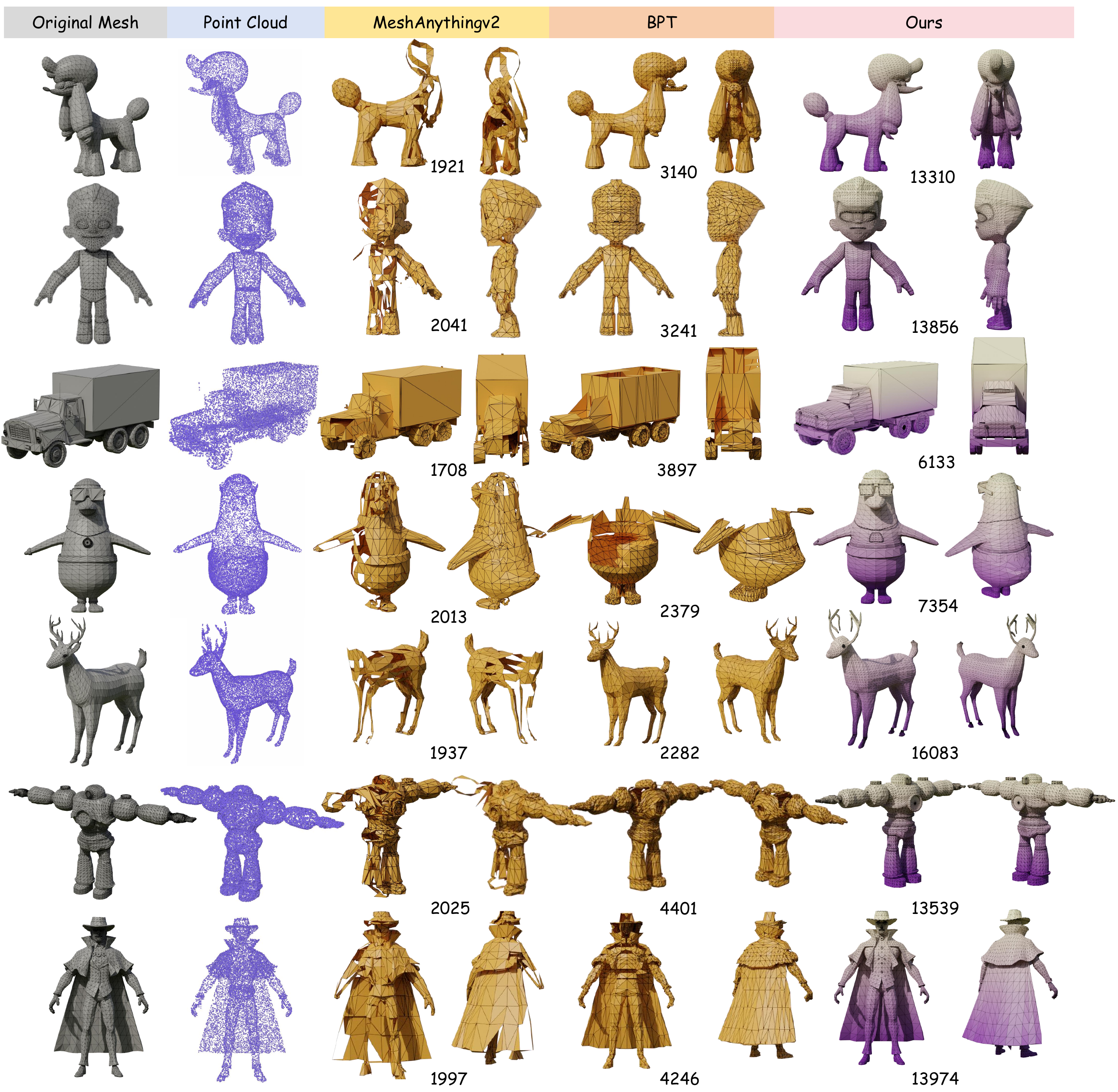}
    \vspace{-0.03\textheight}
    \caption{\textbf{Qualitative comparison on point cloud conditioned generation between DeepMesh and baselines.} DeepMesh outperforms baselines in both generated geometry and preservation of fine-grained details. The meshes generated by ours have much more faces than others.}
    \label{fig:compare}
    
\end{figure*}

\subsection{Performance Enhancement by DPO} \label{sec:dpo}

Although our pre-trained model is capable of generating high-quality meshes, it occasionally suffers from inaesthetic appearance and incomplete geometry. To further enhance the results, we employ Direct Preference Optimization (DPO)~\cite{rafailov2023direct} to align the outputs with human preferences. Moreover, we develop a comprehensive annotation pipeline to curate a preference dataset, enhancing the overall quality of the results.

\subsubsection{Score Standard}
Mesh quality is primarily influenced by two factors: geometric integrity and visual appeal. Therefore, we propose a scoring standard for artist-like mesh generation, which comprehensively accounts for these two aspects. Geometric integrity focuses on the completeness and accuracy of the generated mesh. We employ 3D metrics such as the Chamfer Distance to measure the similarity between the generated mesh and its corresponding ground truth. A lower Chamfer Distance indicates higher fidelity and a more complete geometric representation. On the other hand, visual appeal evaluates the aesthetic qualities of the mesh, including regular wireframes and surface details. Since there exists no score model to assess the visual quality of meshes, we recruit volunteers to compare different meshes and decide which is more visually attractive based on their subjective preferences. This method of gathering human feedback captures aesthetic judgments that conventional metrics might overlook.

\subsubsection{Preference Pair Construction}
We employ our proposed scoring standard to construct a dataset of preference pairs. For each input point cloud, our model generates two distinct meshes and a preference pair is selected. Specifically, we first apply the Chamfer Distance metric to assess the geometric completeness of the generated meshes. If both meshes exhibit a Chamfer Distance above a predefined threshold, they are discarded. In cases where one mesh showcases high geometric fidelity while the other suffers from deficiencies, the superior mesh is designated as the preferred choice. Finally, if both meshes meet the geometric criteria, volunteers are engaged to express their aesthetic preferences. Their judgments help determine the chosen response, ensuring that it aligns with human-like preferences. Figure \ref{fig:dpo_data} presents some examples of our collected data pairs, each distinguished by geometry and appearance appeal. We totally collect 5,000 preference pairs to support the post-training of DPO.

\subsubsection{Direct Preference Optimization}
DPO ~\cite{rafailov2023direct} is used to align generative models with human preferences. By training on pairs of generated samples with positive ($y^+$) and negative labels ($y^-$), the model learns to generate positive samples with higher probability. The objective function of DPO is formulated as:
\begin{equation}
\begin{aligned}
\mathcal{L}_{\mathrm{DPO}}\left(\pi_\theta ; \pi_{\mathrm{ref}}\right)
=& -\mathbb{E}_{\left(c, y^+, y^-\right) \sim \mathcal{D}} \Biggl[
\log \sigma\Bigl(
\beta \log \frac{\pi_\theta\left(y^+ \mid c\right)}{\pi_{\mathrm{ref}}\left(y^+ \mid c\right)} \\
&\quad - \beta \log \frac{\pi_\theta\left(y^- \mid c\right)}{\pi_{\mathrm{ref}}\left(y^- \mid c\right)}
\Bigr)
\Biggr].
\end{aligned}
\end{equation}
where $\beta$ is coefficient that balance preferred and dispreferred terms. We post-train our model using the constructed preference pairs dataset with the above loss function to align its outputs with both geometric fidelity and aesthetic appeal. Additionally, to maintain training efficiency, we also adopt the same truncated training strategy used in the pre-training stage to handle long token sequences, which are generated by the high-poly meshes in dataset.

\begin{figure}[h]
    \centering
    \includegraphics[width=\linewidth]{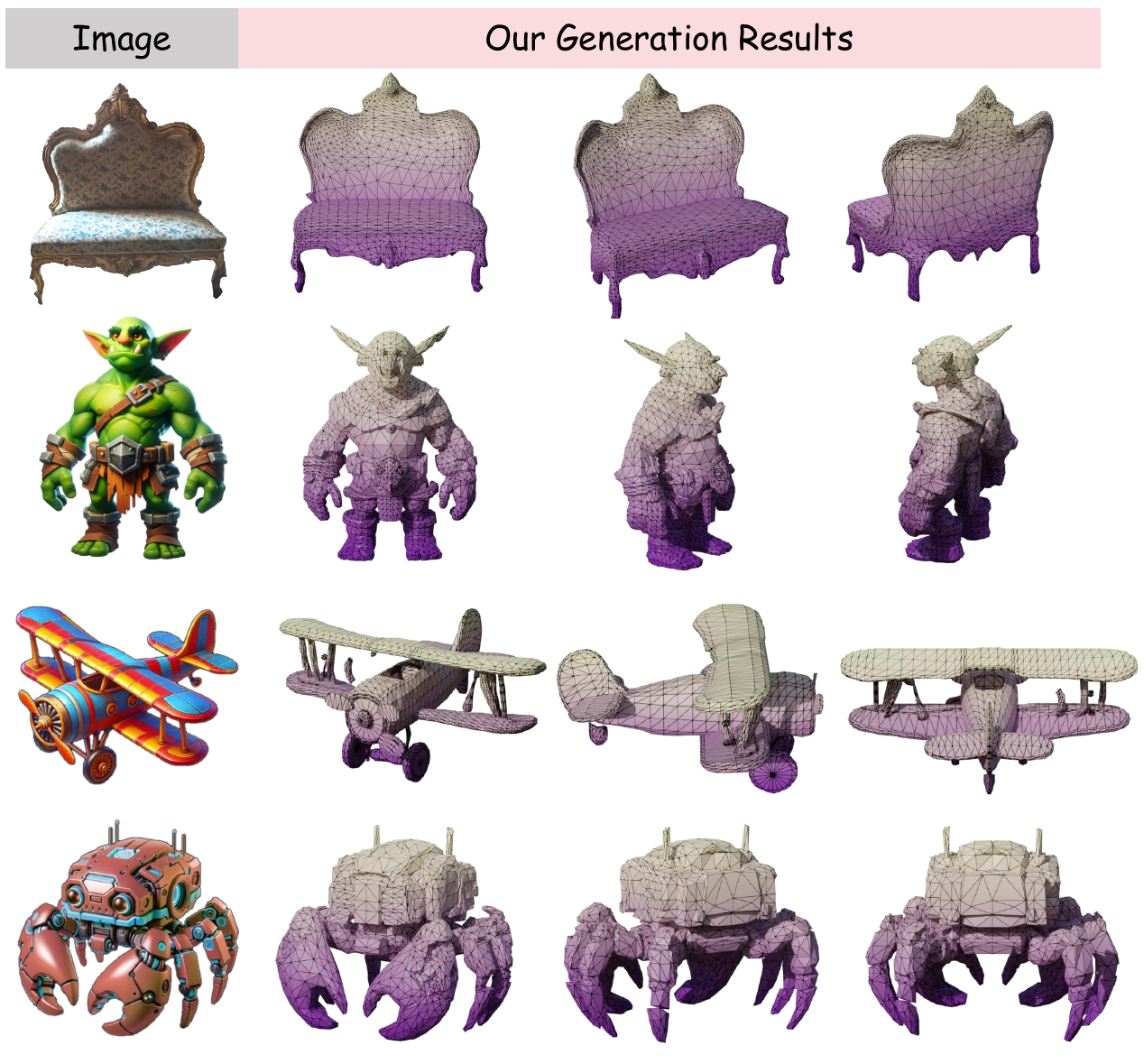}
    \vspace{-0.03\textheight}
    \caption{\textbf{Image-conditioned generation results of our method.} Our method can generate high-fidelity meshes aligned with the input images. }
    \label{fig:image_compare}
\end{figure}

\section{Experiments}
\subsection{Implementation Details}
Our model is trained on the mixture of ShapeNetV2~\cite{chang2015shapenet}, ABO~\cite{collins2022abo}, HSSD~\cite{khanna2024habitat}, Objaverse~\cite{deitke2023objaverse}, Objaverse-XL~\cite{deitke2023objaversexl} and licensed data. To yield better generalization ability, we randomly rotate the meshes with with degrees from ($0^\circ$, $90^\circ$, $180^\circ$, $270^\circ$). For each mesh, we sample $20k$ points and randomly select $16,384$ points as the condition. Our model is pre-trained on 128 NVIDIA A800 GPUs for 4 days, with a cosine learning rate scheduler from $1e-4$ to $1e-5$. For the post-training stage with DPO, we fine-tune our model with a learning rate of $1e-5$ for $10$ epoch. The model’s truncated context length is set to $9k$ tokens. During mesh generation, we use probabilistic sampling with a temperature of $0.5$ for stability.  More implementation details can be seen in supplementary material.

\subsection{Qualitative Results}
\subsubsection{Point-cloud Conditioned}
For point cloud conditioned generation, we compare our results with the latest open-source artist-like mesh generation in the point cloud conditioned generation, such as MeshAnythingv2~\cite{chen2024meshanythingv2} and BPT~\cite{weng2024scaling}. As illustrated in Fig \ref{fig:compare}, the baselines fails to capture fine topological details, suffering from surface holes and missing components. In contrast, our method generates aesthetically appealing artist-like meshes that preserve the geometric details. This is because we improve the tokenization algorithm for high-resolution meshes and further align the model with human preference with DPO. Moreover, while the baselines generate only simple meshes with a few faces, our approach is capable of producing high-quality meshes with much more faces, which benefits from our adopted truncated training strategy.
\subsubsection{Image Conditioned}
For image-conditioned generation, we first utilize TRELLIS~\cite{xiang2024structured} for image-to-3D generation. Then we sample point clouds from the meshes to conduct point-cloud conditioned generation. Figure \ref{fig:image_compare} demonstrates our high-quality generated outputs.

\begin{figure}[h]
    \centering
    \includegraphics[width=\linewidth]{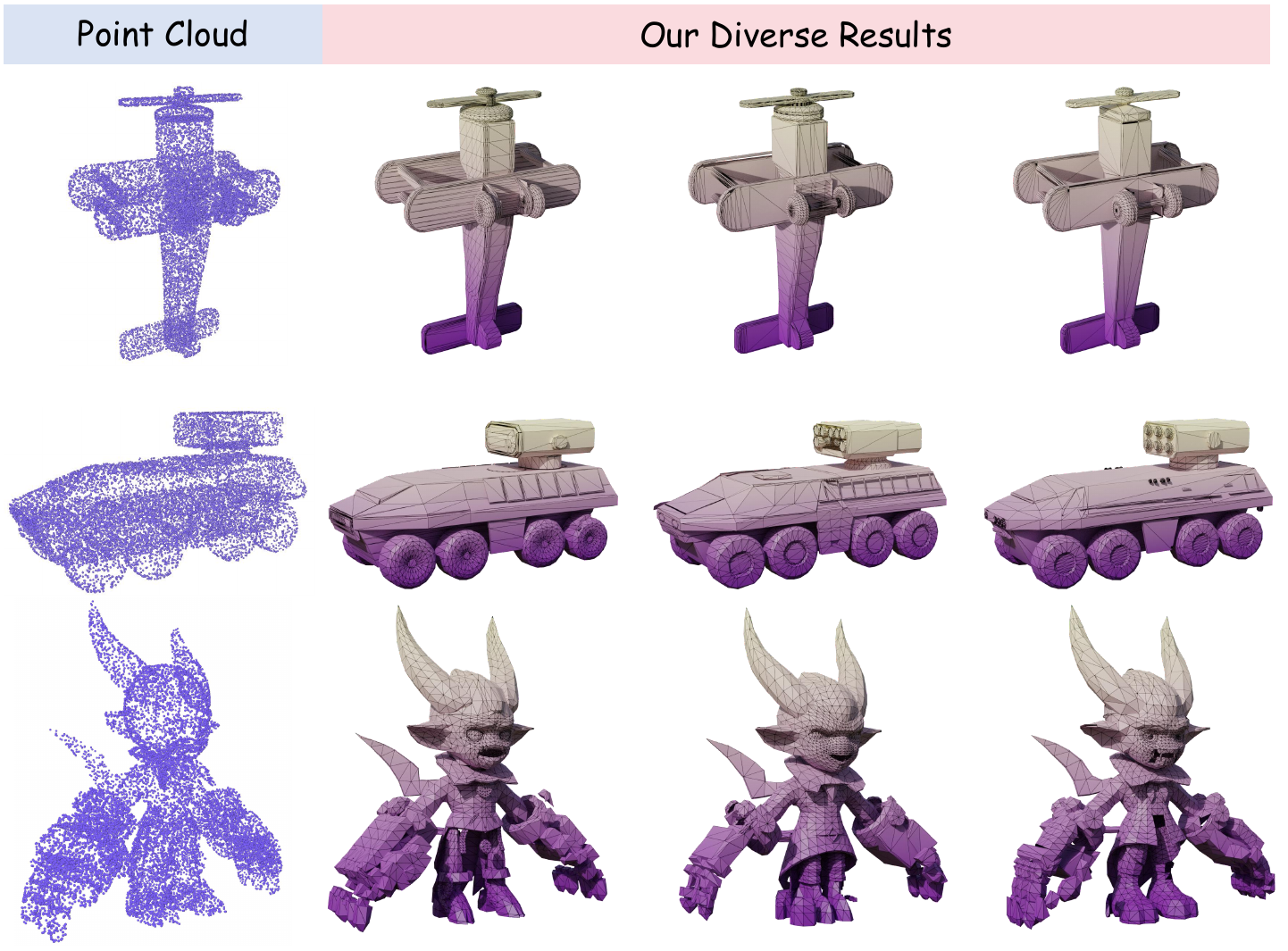}
    \vspace{-0.03\textheight}
    \caption{\textbf{Diversity of generations.} DeepMesh can generate meshes with diverse appearance given the same point cloud.}
    \label{fig:diverse}
\end{figure}
\begin{figure*}[th]
    \centering
    \includegraphics[width=\linewidth]{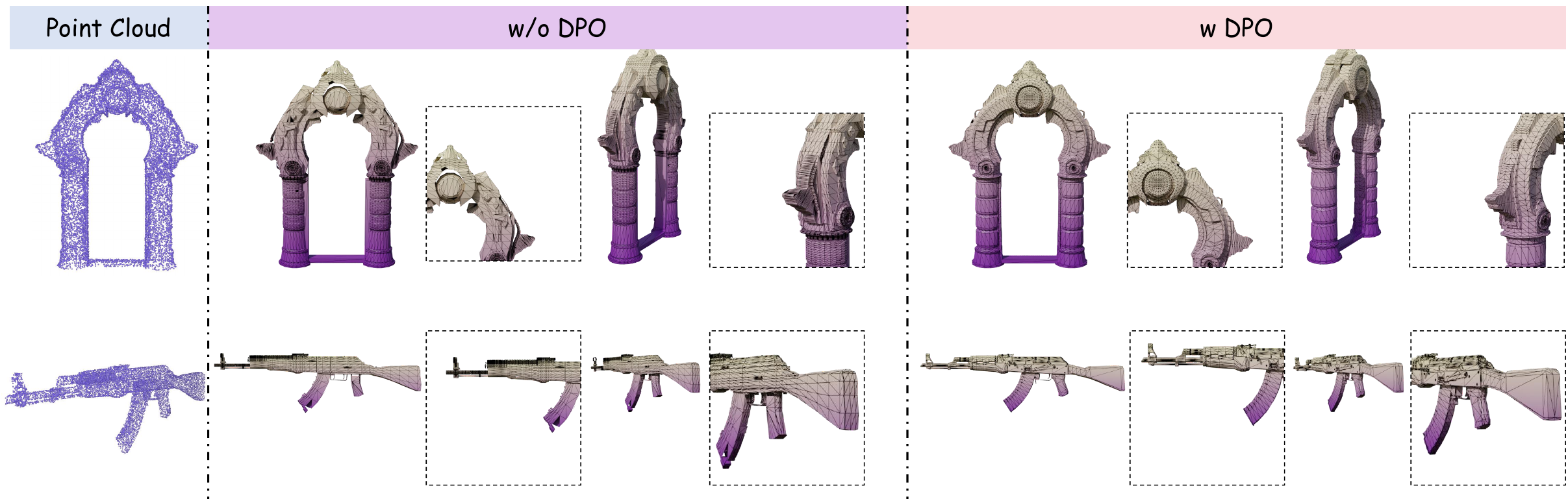}
    \vspace{-0.03\textheight}
    \caption{\textbf{Ablation study on the effectiveness of DPO.} We can observe that while both approaches yield excellent geometry, the results generated using DPO are more visually appealing.}
    \label{fig:dpo}
\end{figure*}
\subsubsection{Diversity}
We evaluate the diversity of generated meshes by providing the same point clouds repeatedly and observe the variations in the meshes. Figure \ref{fig:diverse} shows our model generates a variety of distinct meshes that are consistent with the input point cloud, highlighting the ability to produce creative high-fidelity outputs with diverse appearance. This diversity is crucial for applications that require multiple design options and variations.

\subsection{Quantitative Results}
We compared our point cloud-conditioned results with MeshAnythingv2 and BPT on a test dataset of 100 meshes generated from ~\cite{xiang2024structured}. Similar to~\cite{weng2024scaling,wang2025nautilus}, we uniformly sample 1,024 point clouds from the surfaces of ground truth and generated mesh and compute the Chamfer Distance (C.Dist.) and Hausdorff Distance (H.Dist.) between them. As shown in Table \ref{tab:geometry}, our method outperforms all of the baselines in geometry similarity. In addition, we also conduct a user study to assess the subjective visual appeal of the generated meshes. Volunteers are asked to compare our results with those produced by baselines. It is also can be found that the generation results of ours are most preferred.
\begin{table}
  \centering
  \begin{tabular}{@{}lccc@{}}
    \toprule
    Metrics & C.Dist. $\downarrow$ & H.Dist. $\downarrow$ & User Study $\uparrow$ \\
    \midrule
    MeshAnythingv2~\cite{chen2024meshanythingv2} & 0.1249 & 0.2991  & 10\% \\
    BPT~\cite{weng2024scaling}  & 0.1425 & 0.2796  & 19\% \\
    Ours w/o DPO & 0.1001 & 0.1861 & 34\% \\
    Ours w DPO & \textbf{0.0884} & \textbf{0.1708} &\textbf{37\%}\\
   
    \bottomrule
  \end{tabular}
  \caption{\textbf{Quantitative comparison with other baselines.} Our method outperforms other baselines in generated geometry and visual quality.  }
  \label{tab:geometry}
\end{table}

\subsection{Ablation Study}

\begin{table} 
  \centering
  \begin{tabular}{@{}lcccc@{}}
    \toprule
    Metrics & AMT & EdgeRunner & BPT~  & Ours \\
    \midrule
    Comp Ratio $\downarrow$ & 0.46 & 0.47 & \textbf{0.26} & 0.28 \\
    Vocal Size $\downarrow$ & \textbf{512} & \textbf{512} & 40960 & 4736 \\
    Time (s) $\downarrow$ & 816 & - & 540 & \textbf{480} \\
   
    \bottomrule
  \end{tabular}
  \caption{\textbf{Quantitative comparison with other tokenization algorithms.} Our improved tokenization algorithm achieves a low compression ratio, a small vocabulary size, and the highest computational efficiency, making it both compact and highly efficient for mesh processing. }
  \label{tab:tokenization}
\end{table}

\subsubsection{Tokenization Algorithm} \label{exp:token}
 We compare our tokenization algorithm for 512-resolution meshes with Adjacent Mesh Tokenization (AMT)~\cite{chen2024meshanythingv2}, EdgeRunner~\cite{tang2024edgerunner}, and BPT~\cite{weng2024scaling}. First, we assess the compression ratio, defined as the reduction in sequence length relative to the vanilla representation (which corresponds to nine times the number of faces). A lower compression ratio indicates a more compact representation, which improves storage and computational efficiency. Second, we calculate the vocabulary size, which impacts the complexity of model training. Larger vocabularies indicate greater memory storage and training difficulty. Moreover, we evaluate the training time of different methods on a test dataset comprising 80 meshes, each with a face count of 20k. As shown in Table~\ref{tab:tokenization},  our tokenization method achieves a balanced trade-off between compression ratio and vocabulary size while outperforming all baselines in computational efficiency.

\subsubsection{DPO Post-training}
We collected human preference pairs and fine-tuned our pre-trained model using DPO to enhance its capability to generate meshes with superior geometry and aesthetics. To validate the effect of DPO, we compared the outputs of the post-trained model with those of the pre-trained model. Figure \ref{fig:dpo} indicates that the post-trained model exhibits a clear advantage over the pre-trained one. This suggests the importance of learning from preference pairs, which reduces the likelihood of generating suboptimal outputs. Additionally, quantitative evaluations presented in Table \ref{tab:geometry} demonstrate that DPO-enhanced results have the most similarity with the ground truth and are most preferred by users.

\section{Conclusion}
We introduce DeepMesh, a novel approach that generates artist-like meshes with reinforcement learning. By improving the tokenization algorithm for high-resolution meshes, we preserve intricate details of high-poly meshes while achieving significant sequence compression. We also introduce several pre-training strategies including data curation and data packaging to boost training efficiency. In addtion, by aligning results with human preferences using DPO~\cite{rafailov2023direct}, we refine both topology and visual quality of the generated meshes. The extensive experiments demonstrate that DeepMesh outperforms existing methods across various metrics, enabling the creation of meshes with geometric complexity and details. 

\nocite{zhao2025riflex}
\nocite{liu2024reconxreconstructscenesparse}
{
    \small
    \bibliographystyle{ieeenat_fullname}
    \bibliography{main}
}

\begin{figure*}[t]
    \centering
    \includegraphics[width=\linewidth]{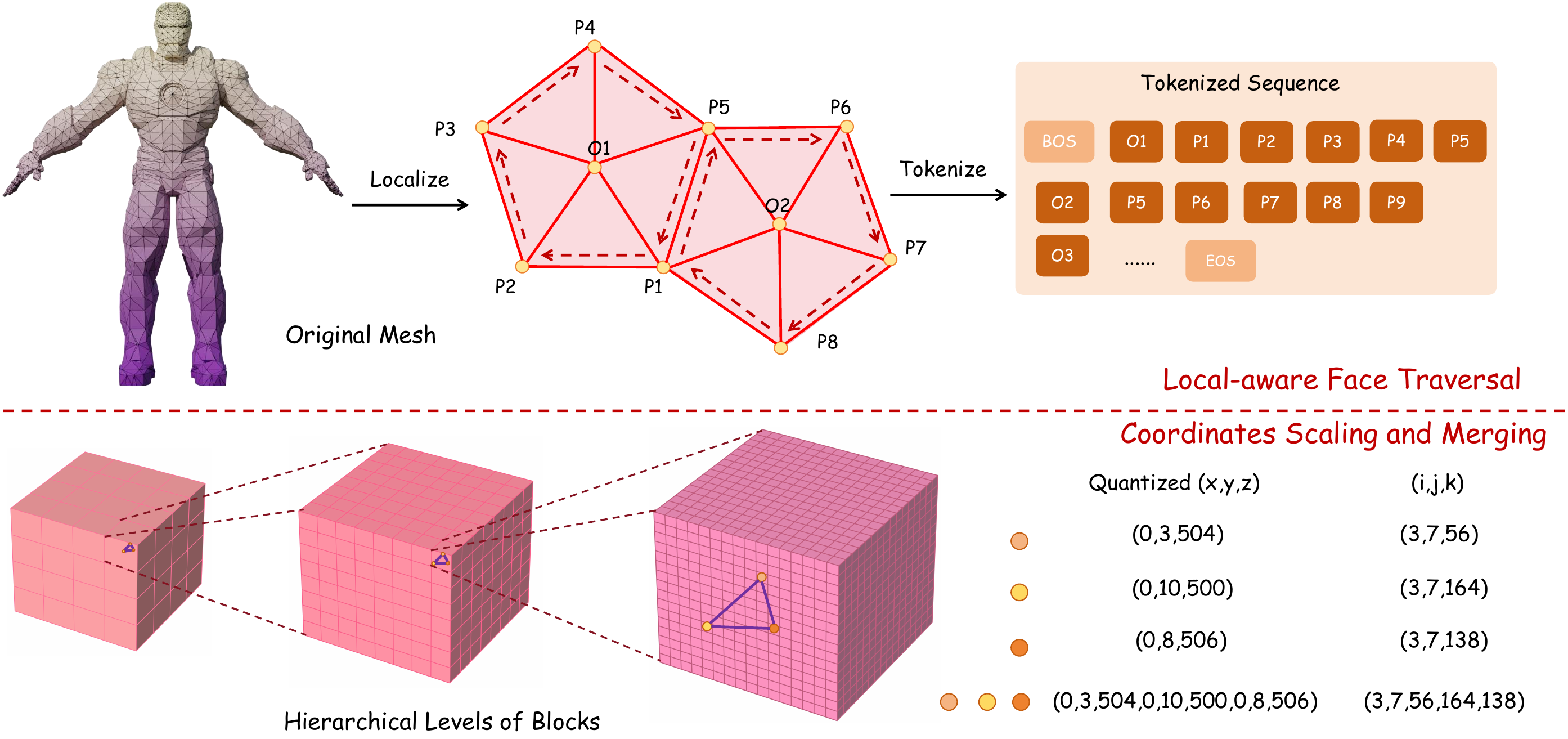}
    \vspace{-0.03\textheight}
    \caption{\textbf{Details of our tokenization algorithm.} We first traverse mesh faces by dividing them into patches according to their connectivity and quantize each vertex of faces into $r$ bins (in our setting $r=512$).Then we partition the whole coordinate system into three hierarchical levels of blocks and index the quantized coordinates as offsets within each block. We merge the index of neighbor vertices if they have the identical values.}
    \label{fig:tokenization_framework}
\end{figure*}
\newpage
\appendix

\section{Details of Tokenization Algorithm}
In this section, we detail our improved tokenization algorithm. As illustrated in Figure \ref{fig:tokenization_framework}, we first traverse mesh faces to reduce redundancy in the vanilla mesh representation. Specifically, we divide mesh faces into multiple local patches according to their connectivity similar to ~\cite{weng2024scaling}. Each local patch is formed by grouping a central vertex $O$ with its adjacent vertices $P_{1:n}$, which are organized based on their connectivity order:
\begin{equation}
    L_O = (O, P_1, P_2, \cdots ,P_n)
\end{equation}
This organization helps maintain local mesh connectivity by explicitly encoding edge-sharing relationships between adjacent faces. To find each center vertex, we begin by sorting all the unvisited faces. Next, we select the first unvisited face and choose the vertex connected to the most unvisited faces as the center. Then, we iteratively traverse the neighboring vertices within the center's unvisited faces, expanding the local patch by adding adjacent vertices that connect to the current patch. Once the patch is complete, we mark all its faces as visited. We repeat the process above until every face is visited.

After the local-wise face traversal, we normalize and quantize each vertex in faces and flatten them in $XYZ$ order. With a resolution of $r$, coordinates of each vertex are quantized into $[0,r-1]$ (in our setting, $r=512$).  The coordinates of all vertices are then concatenated to form a complete sequence of tokens. To further reduce the sequence length, we partition the whole coordinate system into three hierarchical levels of blocks and index the quantized coordinates as offsets within each block, as shown in Figure \ref{fig:tokenization_framework}. The volume of each block is $A$, $B$ and $C$ respectively. In our setting, $A=4$, $B=8$ and $C=16$. We scale quantized Cartesian coordinate $(x,y,x)$ of each vertex into $(i,j,k)$ by:
\begin{equation}
\begin{aligned}
 i=&\left(x \mid B\cdot C\right) \cdot A^2+\left(y \mid B\cdot C\right) \cdot A+(z \mid B\cdot C) \\
 j = &(x \% B \cdot C \mid C) \cdot B^2 +(y \% B \cdot C \mid C) \cdot B \\
 &+ (z \% B \cdot C \mid C)\\
 k=&\left(x \% C\right) \cdot C^2+\left(y \% C\right) \cdot C+z \% C
\end{aligned}
\end{equation}
As the coordinates are sorted, it is common for neighbor vertices to share the same offset in block. Therefore,  we merge the adjacent $(i,j,k)$ if they have the identical values to save more length. Specifically, for vertices $v_i,v_2 ,\cdots, v_n$, the sequence of their coordinate representation can be simplified as follows:
\begin{equation}
\begin{aligned}
    (v_1, v_2, \cdots, v_n) = (&i_1, j_1, k_1, i_1, j_1, k_2, \cdots, i_1,j_2, k_{s+1})\\=(&i_1, j_1, k_1, k_2, \cdots ,k_s, j_{2}, k_{s+1}, \cdots, k_n)
\end{aligned}
\end{equation}

To distinguish different patches, we extend the vocabulary size of $i$ and $j$ for each center vertex in patches. This design eliminates the need for special tokens to separate adjacent local patches, thereby avoiding unnecessary increases in mesh sequence length.
\begin{figure}[t]
    \centering
    \includegraphics[width=\linewidth]{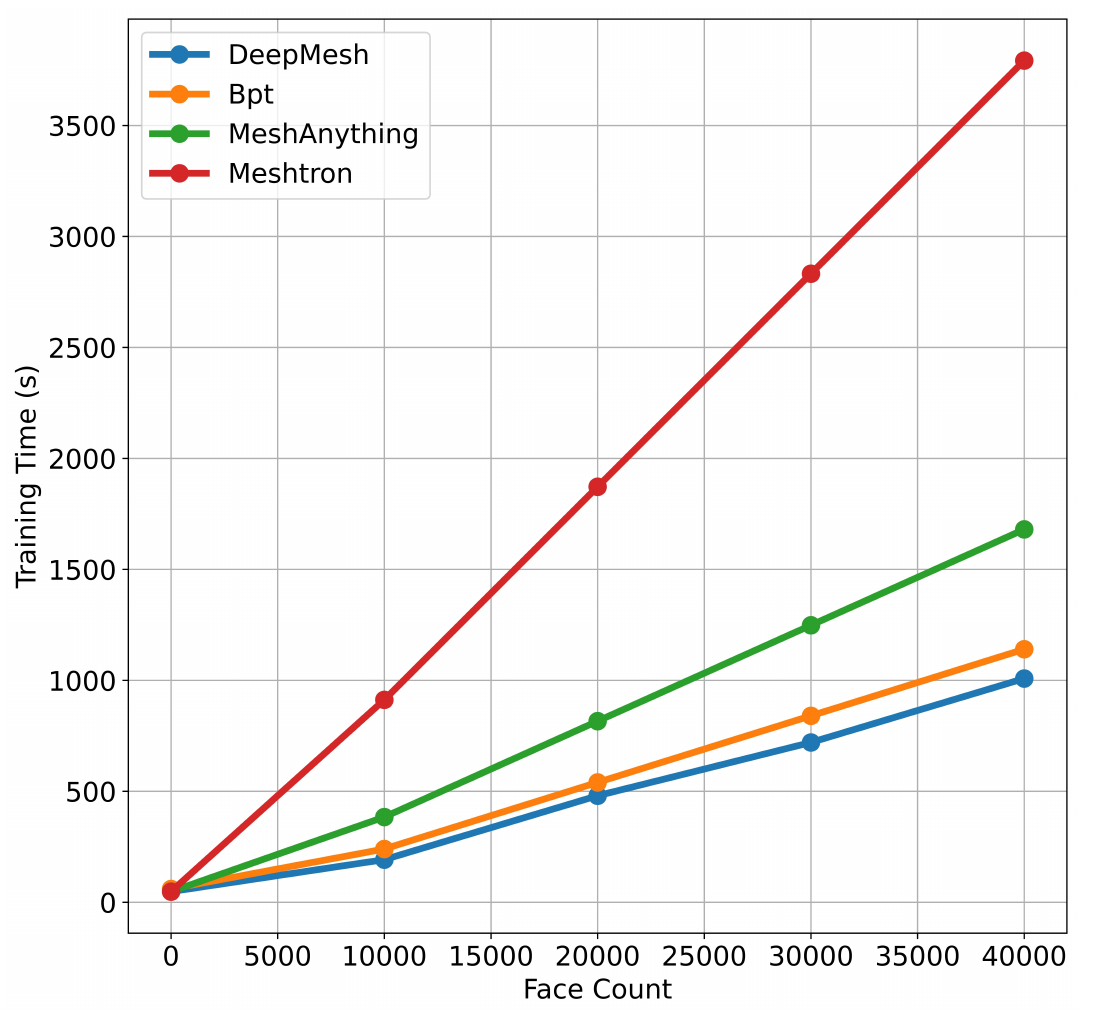}
    \vspace{-0.03\textheight}
    \caption{\textbf{Comparison with other tokenization algorithms in training effciency.} We integrate all tokenization algorithms into our model architecture and train them on a dataset of 80 meshes for each face count category (10K, 20K, 30K, 40K). Our method achieves the fastest training time across all face count categories, demonstrating superior training efficiency.}
    \label{fig:tokenization}
\end{figure}

\section{More Implementation Details}
\subsection{Training Data Filtering Pipeline}
\label{sec:data-filter}
The data in training dataset varies significantly in quality, which may lead to three primary challenges: 1. Unstructured topology that fails to meet the artist-mesh standard. 2. Fragmented data that cannot assemble into complete surfaces. 3. Overly complex structures, such as characters with tangled or messy hair geometry.

To efficiently filter out low-quality data, we propose an data-filtering pipeline comprising the following four stages:

(1) First of all, We remove meshes with a mesh.area metric below 1 to filter out the fragmented data.

(2) Then, We construct a high-quality subset of the training data and perform low-cost pretraining on it to build a baseline model.

(3) Subsequently, the pretrained model is then used to evaluate the remaining data, recording their test losses. Meshes exceeding a predefined loss threshold are flagged and placed on a candidate deletion list.

(4) Finally, The candidate meshes are rendered into images and scored using a pretrained aesthetic assessment model~\cite{xiang2024structured}. We retain the top 20\% of the highest-scoring meshes to ensure that high-quality but complex meshes are not mistakenly removed.

After filtering out the poor-quality data, our dataset size decreases from 800k to approximately 500k, with an average face count of 8k.

\subsection{Preference Pair Constructed Pipeline}
The point clouds in our preference pair dataset come from the training dataset and a manually selected high-quality test dataset. This ensures a diverse  dataset for learning human preference. Since high-poly mesh generation is extremely time-consuming—for example, our full-scale model requires at least 10 minutes to generate a single mesh with over 30K faces—it is crucial to pre-select a candidate list when constructing the DPO dataset to improve efficiency and feasibility. To ensure that the preference data is representative, we filter out overly simple and complex meshes. Specifically, we follow a similar data-filtering approach in Section \ref{sec:data-filter}, removing samples with fewer than 5,000 faces as well as those with extremely high or low test loss. Using the remaining curated data, we then construct our preference pair dataset for post-training. For mesh generation, we use a temperature of 0.5 and generate 2 meshes for each point cloud.

\subsection{More Training Details}
We respectively train a small-scale model and a large-scale model for DeepMesh, with architecture details provided in the Table \ref{tab:details}. We train both of the models for 100k iteration steps to ensure convergence. Moreover, we employ FlashAttention and Zero2 to reduce GPU memory usage.

\subsection{Hourglass Transformers}
Inspired by ~\cite{hao2024meshtron,nawrot2021hierarchical}, we adopt the Hourglass Transformers architecture for efficient training. For hyperparameters, we maintain the settings from ~\cite{hao2024meshtron}. Specifically, the shortening factor is set to 3, while both the downsampling and upsampling layers are used with the Linear layers.

\begin{table}
  \centering
  \begin{tabular}{lcc}
    \toprule
 & Small scale & Large scale \\
 \hline
Parameter count & 500 M &1.1B\\
Batch Size & 9&5\\
Layers &21& 20 \\
Heads& 10& 14\\
$d_{model}$&1280 & 1792\\
$d_{FFN}$&5120 & 7168 \\
Learning rate & $1e-4$ &$1e-4$ \\
LR scheduler& Cosine& Cosine  \\
Weight decay & 0.1 & 0.1  \\
Gradient Clip & 1.0 & 1.0 \\

    \bottomrule
  \end{tabular}
  \caption{Deepmesh's architectural and training details.}
  \label{tab:details}
\end{table}

\section{More Ablation Study}
\subsection{Efficiency of Tokenization }
\label{tab:Ablation About Compression}
We evaluate the computational efficiency of our mesh tokenization algorithm compared to other baselines~\cite{hao2024meshtron,chen2024meshanythingv2,weng2024scaling}. To ensure a fair comparison, we integrate each method's compressed mesh representation into our model while keeping all other parameters unchanged, as detailed in Table \ref{tab:details}. For training, we use a single GPU and dynamically adjust the batch size to fully utilize available memory. We test on a dataset of 80 meshes for each face count category: 10K, 20K, 30K, and 40K faces. As shown in Figure ~\ref{fig:tokenization}, our method consistently exhibits lowest training time across all face count categories, and achieves the best training efficiency.

\subsection{Data Curation}
\begin{figure}[t]
    \centering
    \begin{subfigure}[b]{\linewidth}
        \includegraphics[width=\linewidth]{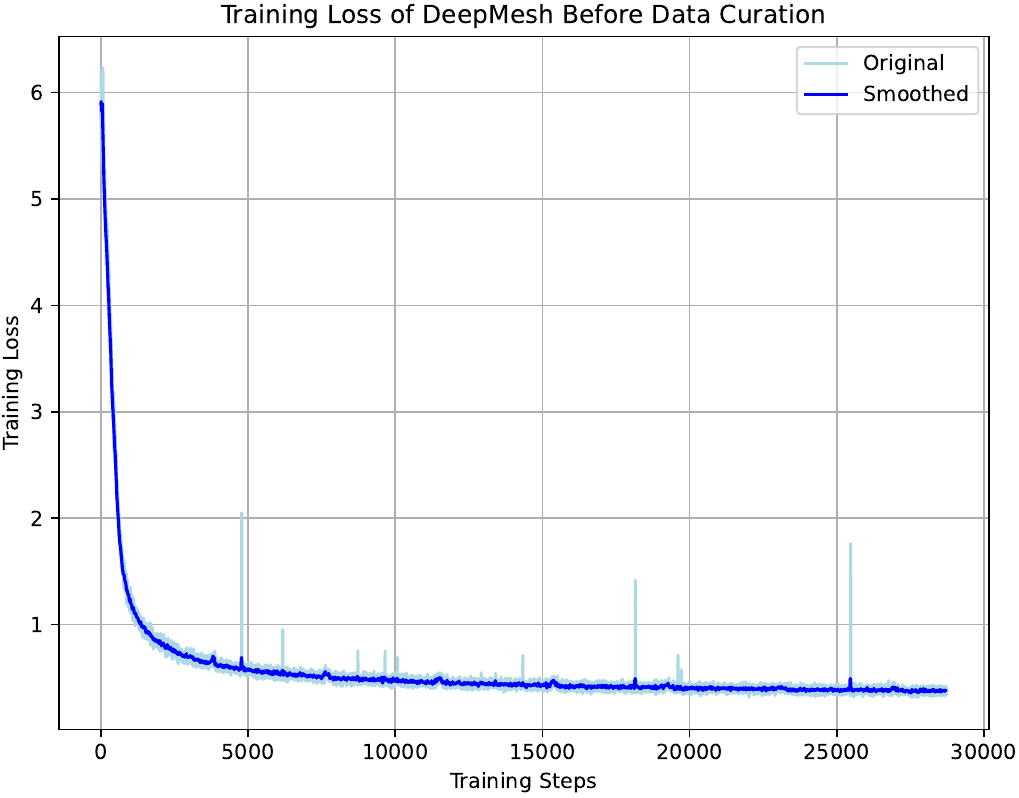}
        \caption{Before data curation}
        \label{fig:loss_before}
    \end{subfigure}
    \hfill
    \begin{subfigure}[b]{\linewidth}
        \includegraphics[width=\linewidth]{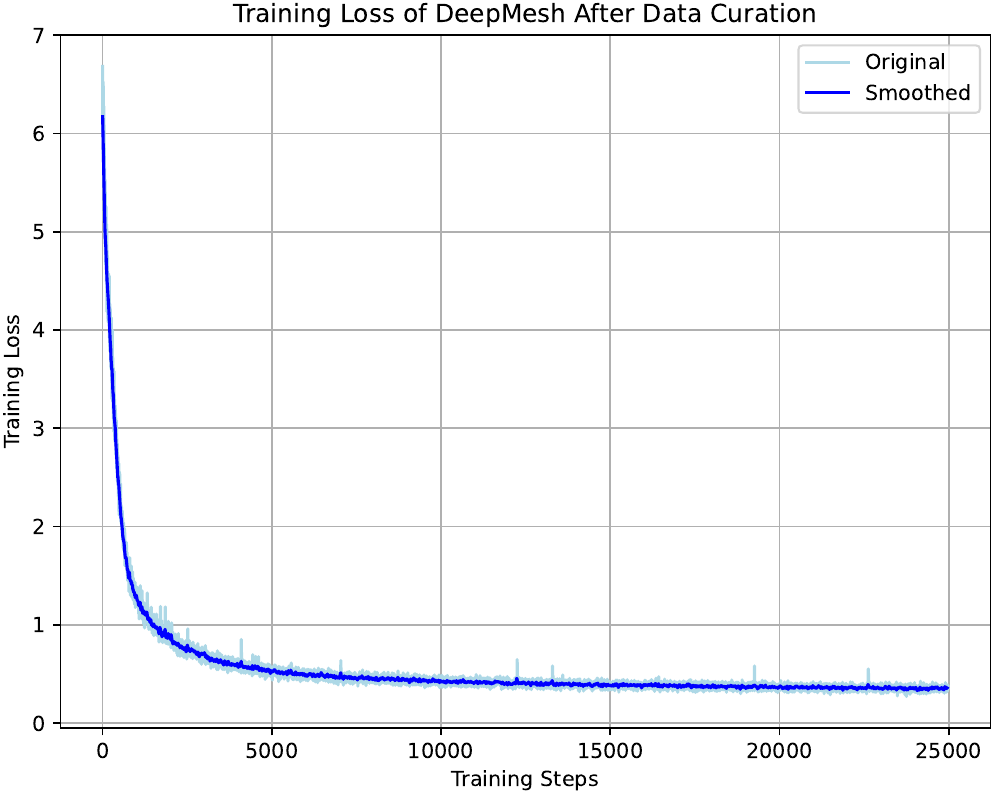}
        \caption{After data curation}
        \label{fig:loss_after}
    \end{subfigure}
    \caption{\textbf{Training loss before and after data curation.} Before data curation, we observe frequent loss spikes. After data curation, pre‑training becomes significantly more stable.}
    \label{fig:loss}
\end{figure}

During the initial stages of training, we observe frequent spikes in the loss curve, as illustrated in Figure \ref{fig:loss}. This suggests that certain training samples lead to irregular loss values, potentially disrupting the learning process. To address this issue, we apply the data filtering strategy outlined in Section \ref{sec:data-filter}, removing low-quality samples to ensure stable training. This filtering process can mitigate the inconsistencies caused by poor mesh structures. The impact of this curation is reflected in the improved training loss curve, also shown in Figure \ref{fig:loss}.

\section{Limitations and Future Work}
Although DeepMesh demonstrates impressive mesh generation capabilities, there are several limitations to address in future work. First, Tthe generation quality of DeepMesh is constrained by the low-level features of point cloud conditioning. As a result, it struggles to recover fine-grained details present in the original meshes. To address this, future improvements could focus on enhancing the point cloud encoder or integrating salient point sampling techniques, such as those proposed in ~\cite{chen2024dora}. Also, DeepMesh is trained on a limited number of 3D data. We believe incorporating more datasets could further enrich the generated results. Additionally, we use only a 1B model due to limited computational resources. We believe that using a larger scale model would further improve the generation quality.
\section{More Results}
\begin{figure*}[th]
    \centering
    \includegraphics[width=\linewidth]{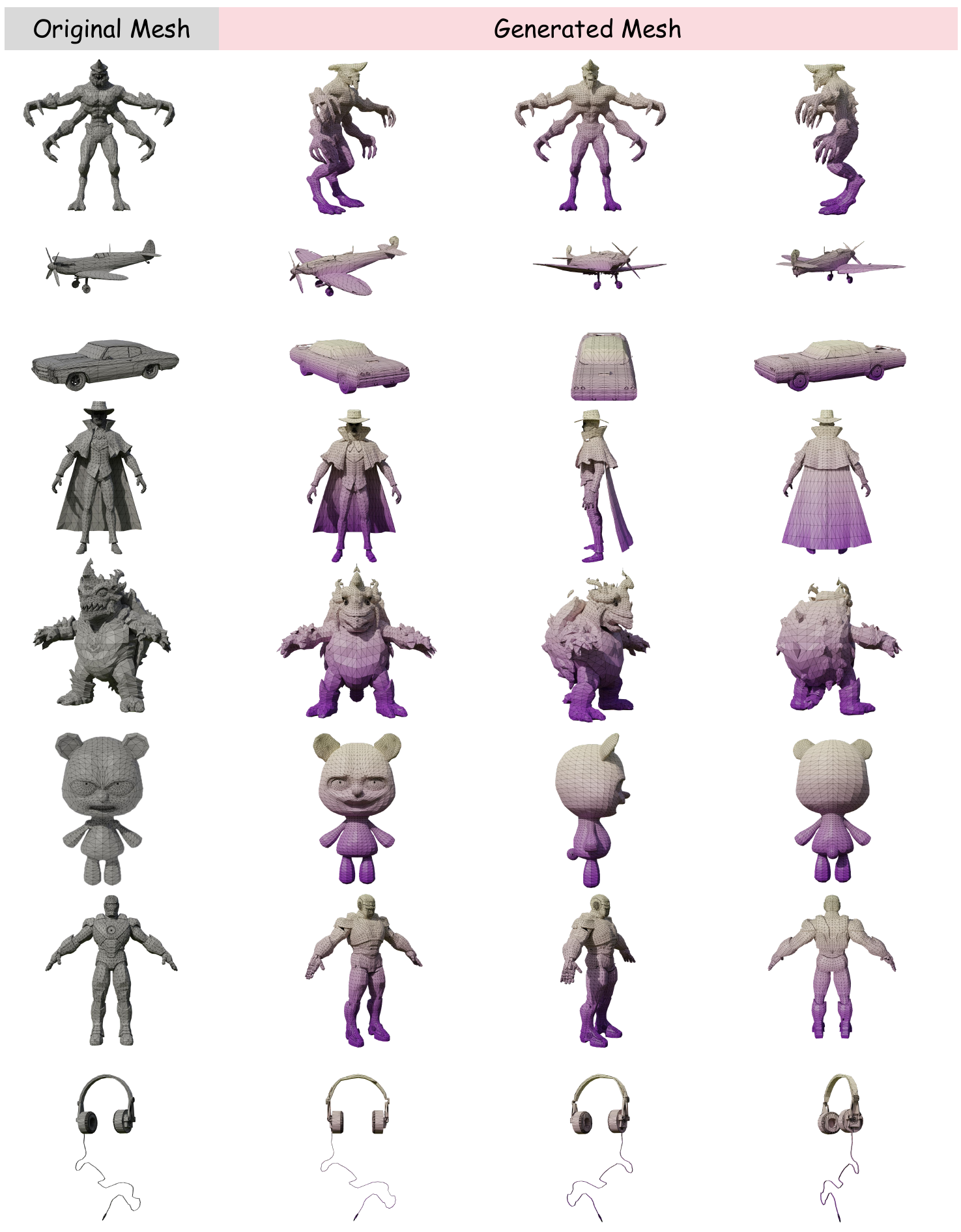}
    \vspace{-0.03\textheight}
    \caption{\textbf{More results of DeepMesh.} We present more high-fidelity results generated by our method.}
    \label{fig:more1}
\end{figure*}

\begin{figure*}[th]
    \centering
    \includegraphics[width=\linewidth]{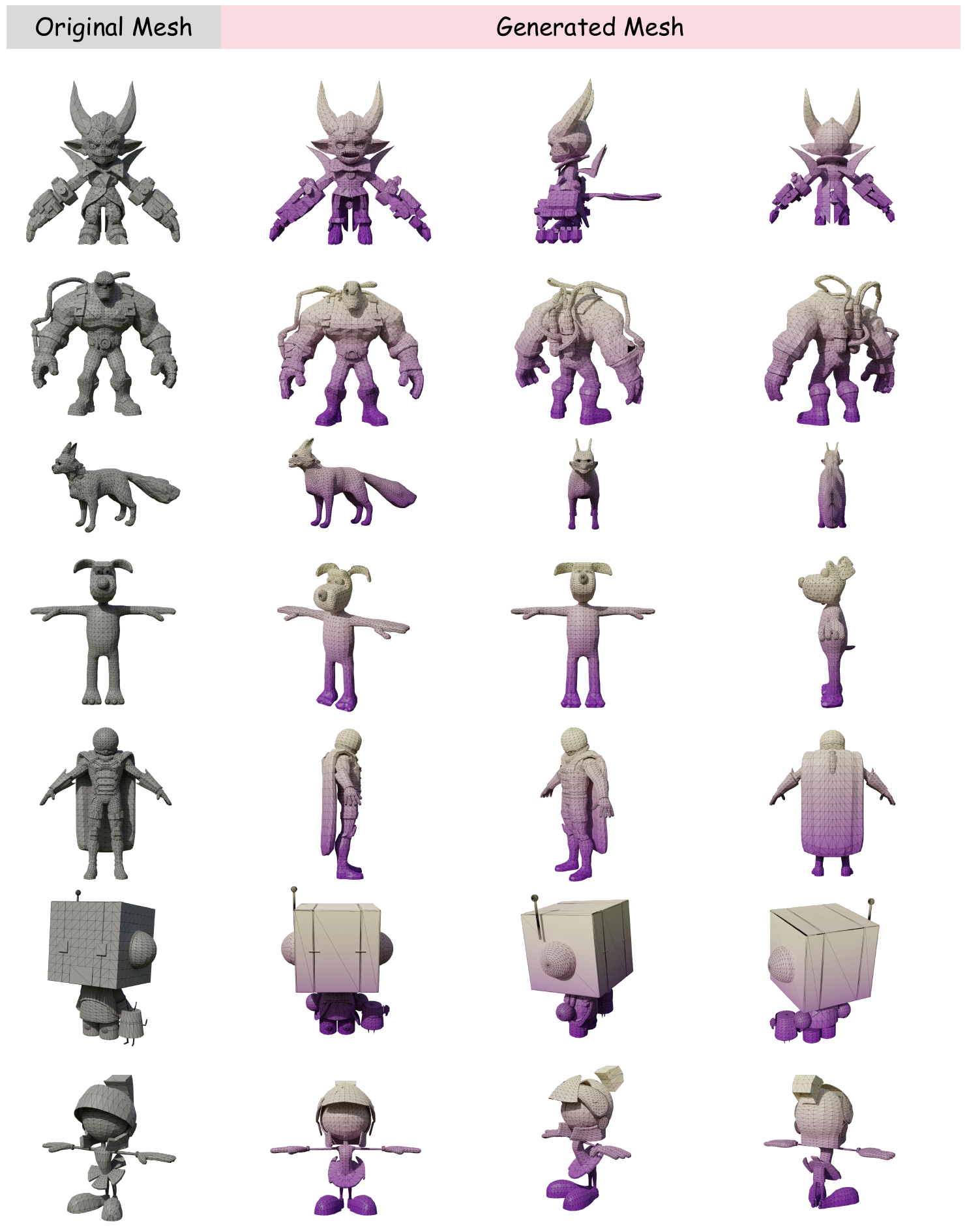}
    \vspace{-0.03\textheight}
    \caption{\textbf{More results of DeepMesh.} We present more high-fidelity results generated by our method.}
    \label{fig:more2}
\end{figure*}

\begin{figure*}[th]
    \centering
    \includegraphics[width=\linewidth]{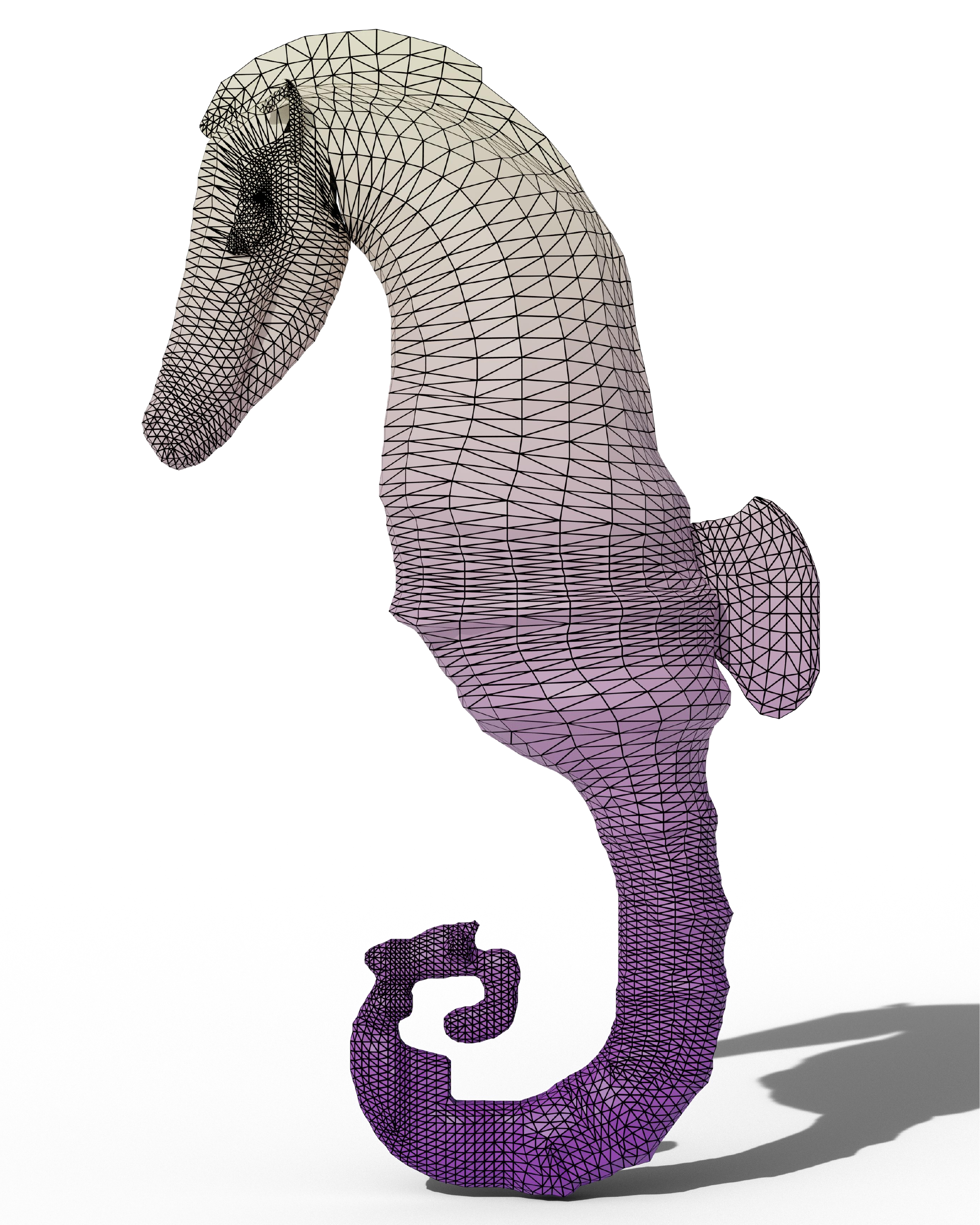}
    \vspace{-0.03\textheight}
    \caption{\textbf{High resolution results of our generated meshes.}}
    \label{fig:haima}
\end{figure*}

\begin{figure*}[th]
    \centering
    \includegraphics[width=1\linewidth]{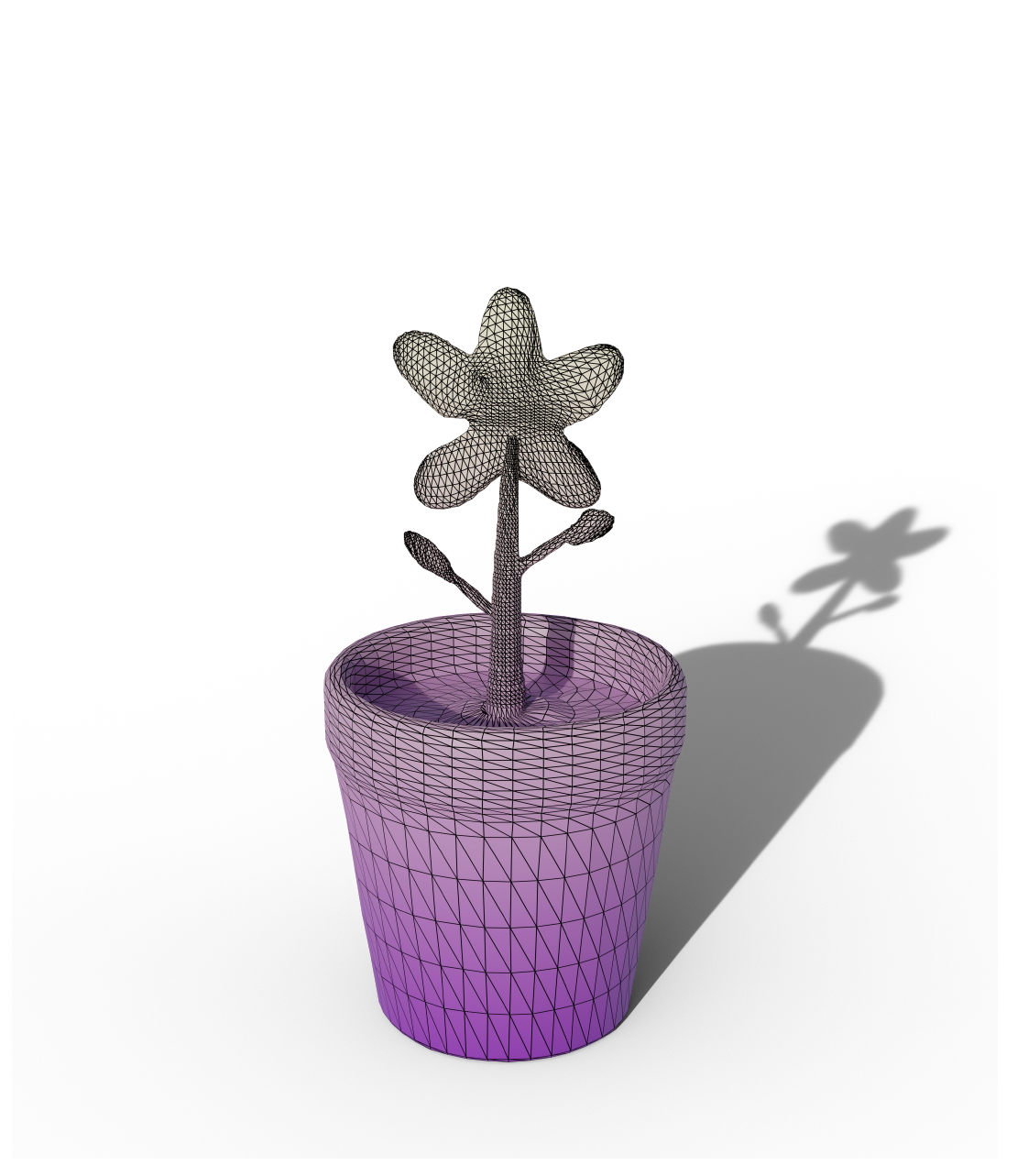}
    \vspace{-0.03\textheight}
    \caption{\textbf{High resolution results of our generated meshes.}}
\label{fig:flower}
\end{figure*}

\begin{figure*}[th]
    \centering
    \includegraphics[width=1\linewidth]{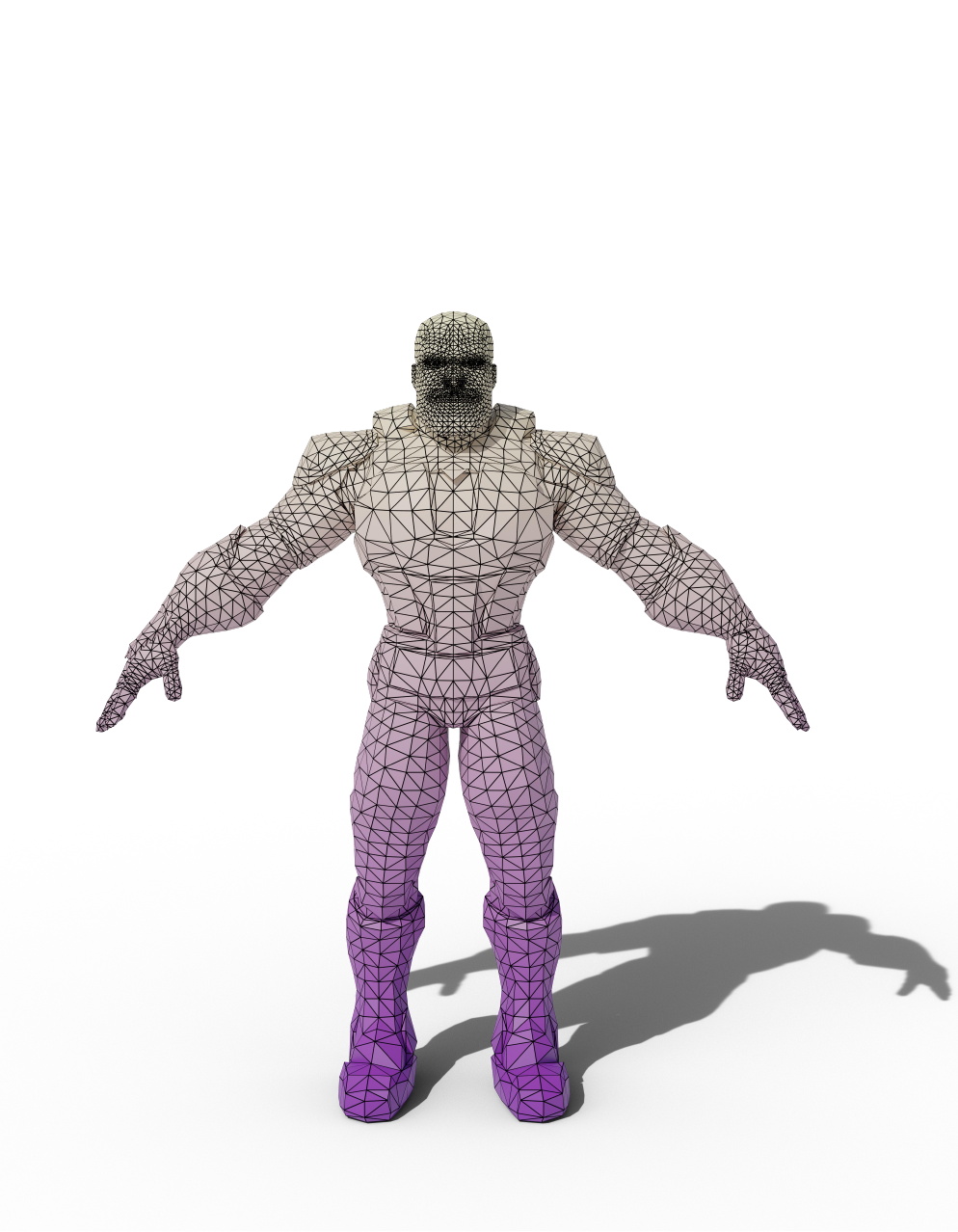}
    \vspace{-0.03\textheight}
    \caption{\textbf{High resolution results of our generated meshes.}}
    \label{fig:man}
\end{figure*}

We provide more visualization results respectively in Figure \ref{fig:more1} and Figure \ref{fig:more2}. Additionally, we select specific cases and present their high-resolution renderings in Figure \ref{fig:haima},\ref{fig:flower} and \ref{fig:man} to see their finer details.

\end{document}